\documentclass[journal]{IEEEtran}
\usepackage{graphicx}
\usepackage{multirow}
\usepackage{booktabs}
\usepackage{amsmath}
\usepackage{amssymb}
\usepackage[switch]{lineno}
\usepackage{amsfonts}
\usepackage{pifont}%
\newcommand{\cmark}{\ding{51}}%
\newcommand{\xmark}{\ding{55}}%
\usepackage{colortbl}
\usepackage{color,xcolor}
\usepackage{flushend}
\usepackage[colorlinks,
linkcolor=red,     
urlcolor=purple,
citecolor=blue,     
]{hyperref}
\usepackage{xcolor, soul}
\definecolor{shadecolor}{rgb}{0.92,0.92,0.92}  
\usepackage{framed}  
\usepackage{cite}  
\ifCLASSINFOpdf
\else
\fi

\usepackage{xcolor}
\usepackage{soul}
\begin{document}
	
	\definecolor{Seashell}{RGB}{250, 250, 0} 
	\definecolor{Firebrick4}{RGB}{0, 0, 0}
	
	\newcommand{\code}[1]{
		\begingroup
		\sethlcolor{Seashell}
		\textcolor{Firebrick4}{\hl{#1}}
		\endgroup
	}
	
	\definecolor{Seashell_pol}{RGB}{0, 250, 0} 
	\definecolor{Firebrick4_pol}{RGB}{0, 0, 0}
	\def\ie{\textit{i.e.}}
	\def\eg{\textit{e.g.}}
	\def\etal{\textit{et al.}}

	\newcommand{\pol}[1]{
		\begingroup
		\sethlcolor{Seashell}
		\textcolor{Firebrick4}{\hl{#1}}
		\endgroup
	}
	
	%
	\title{Comprehensive Visual Question Answering on Point Clouds through  Compositional Scene Manipulation}
	%
	%
	%
	\author{Xu Yan, Zhihao Yuan, Yuhao Du, Yinghong Liao, Yao Guo, Zhen Li, and Shuguang Cui,~\IEEEmembership{Fellow,~IEEE}
		\thanks{Zhen Li is the corresponding author. }
		\thanks{Xu Yan and Zhihao Yuan have equal contributions.}
		\thanks{Xu Yan, Zhihao Yuan, Yuhao Du, Yinghong Liao, Zhen Li, and Shuguang Cui are with The Future Network of Intelligence Institute, The Chinese University of Hong Kong (Shenzhen), and School of Science and Engineering, The Chinese University of Hong Kong (Shenzhen). Emails:\{\small xuyan1@link., zhihaoyuan@link., yuhaodu@link., yinghongliao@link., lizhen@, shuguangcui@\}cuhk.edu.cn}
		\thanks{Yao Guo is with Institute of Medical Robotics in Shanghai Jiao Tong Univerisity, China. Email:yao.guo@sjtu.edu.cn}
	}
	%
	%

	\markboth{Journal of \LaTeX\ Class Files,~Vol.~X, No.~X, August~XX}%
	{Shell \MakeLowercase{\textit{et al.}}: Bare Demo of IEEEtran.cls for IEEE Journals}
	%



	\maketitle
	
	\begin{abstract}
		Visual Question Answering on 3D Point Cloud (VQA-3D) is an emerging yet challenging field that aims at answering various types of textual questions given an entire point cloud scene.
		To tackle this problem, we propose the \textbf{CLEVR3D}, a large-scale VQA-3D dataset consisting of \textbf{171K} questions from \textbf{8,771} 3D scenes.
		Specifically, we develop a question engine leveraging 3D scene graph structures to generate diverse reasoning questions, covering the questions of objects' attributes (\ie, size, color, and material) and their spatial relationships.
		Through such a manner, we initially generated 44K questions from 1,333 real-world scenes.
		Moreover, a more challenging setup is proposed to remove the confounding bias and adjust the context from a common-sense layout.
		Such a setup requires the network to achieve comprehensive visual understanding when the 3D scene is different from the general co-occurrence context (\eg, chairs always exist with tables).
		To this end, we further introduce the compositional scene manipulation strategy and generate 127K questions from 7,438 augmented 3D scenes, which can improve VQA-3D models for real-world comprehension.
		Built upon the proposed dataset, we baseline several VQA-3D models, where experimental results verify that the CLEVR3D can significantly boost other 3D scene understanding tasks.
        Our code and dataset will be made publicly available at \url{https://github.com/yanx27/CLEVR3D}.
		
	\end{abstract}
	
	
	%
	\IEEEpeerreviewmaketitle
	
	\section{Introduction}
	
	
	\IEEEPARstart{3}{D} scene understanding is a critical task in computer vision, aiming to achieve the perception and interpretation of a scene from 3D data.
	This task demands the fundamental perception capability of recognizing and localizing objects in 3D scenes and the high-level reasoning capacities of capturing objects' context and relationships. 
	%
	
	Traditional studies on 3D scene understanding mainly concentrate on the perception tasks such as instance segmentation~\cite{jiang2020pointgroup}, semantic segmentation~\cite{scannet,s3dis} as well as 3D object detection and classification~\cite{qi2019deep,qi2017pointnet++}.
	These works pay more attention to each individual object but ignore the inter-object contexts and relationships, only utilizing them to improve the per-object recognition.
	Recently, applying natural language to improve the scene understanding has become a hot topic, where 3D visual grounding~\cite{chen2020scanrefer,achlioptas2020referit3d}, 3D dense captioning~\cite{chen2021scan2cap,yuan2022x} and scene graph analysis~\cite{wald2020learning, SGGpoint, wu2021scenegraphfusion} are increasingly studied.
	Compared with the reasoning from 2D images, the reasoning in real-world 3D scenes can avoid the inherent spatial ambiguity in 2D and capture the geometric information and inter-object relationships. 
	%
	Despite effort has been gained on enhancing scene comprehension via spatial representations, current works are still deficient in 3D understanding due to the lack of data.
	
	\begin{figure*}[t]
		\begin{centering}
			\includegraphics[width=1\textwidth]{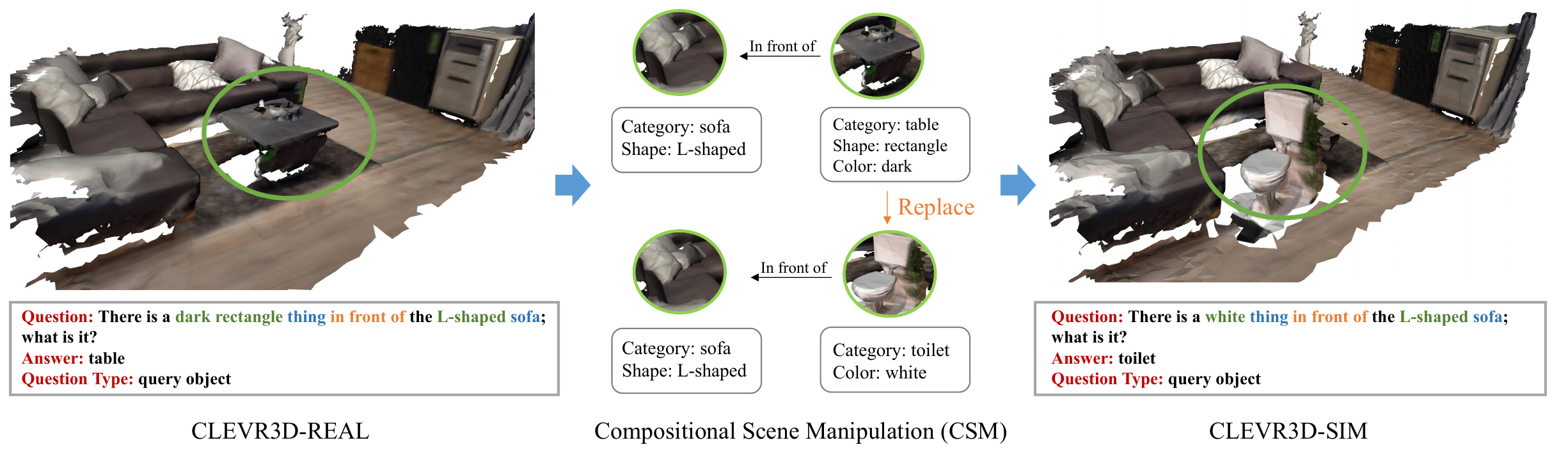}
			\caption{\textbf{Visual Question Answering on 3D Point Cloud (VQA-3D).} In this paper, we introduce a new dataset CLEVR3D, which consists of CLEVR3D-REAL and CLEVR3D-SIM sub-datasets. 
				Selected questions from CLEVR3D-REAL (left) test aspects of visual reasoning in 3D scenes such as counting, object identification, query attribute, and attribute comparison. 
				Each question contains \textbf{\textcolor[rgb]{0,0.3,0.7}{objects}}, \textbf{\textcolor[rgb]{0,0.4,0}{attributes}} and their  \textbf{\textcolor[rgb]{1,0.5,0.3}{ relationship}}.
				CLEVR3D-SIM dataset is obtained through the compositional scene manipulation (CSM) shown in right for common-sense-independent VQA-3D.
			}
			\label{fig:fig1}
		\end{centering}	
		
	\end{figure*}

	This paper introduces a new dataset for Visual Question Answering on 3D Point Cloud (\textbf{VQA-3D}).
	Unlike previous 3D scene understanding tasks only focusing on specific aspects \eg, grounding or captioning, VQA-3D expects the model to answer questions of various types, \ie, verifying object existence, counting, comparison, query object, and query attributes.
	Inspired by well-studied VQA on images~\cite{johnson2017clevr}, our CLEVR3D dataset is generated based on the 3D semantic scene graphs~\cite{wald2020learning}.
	Specifically, we develop a \textit{question engine} by leveraging objects' attributes and inter-object relationships provided by the semantic scene graph of each scene.
	To this end, over 44K questions covering 13 question types on the real-world scenes are obtained.
	%
	Moreover, we investigate a more challenging setup, \ie, common-sense-independent VQA-3D (\textbf{CSI-VQA-3D}).
	Since objects are not alone in the 3D scenes and usually co-occur under specific contexts, the VQA models are susceptible to such confounding priors and thus learn spurious associations.
	%
	For example, it is easy to learn that ``table'' is in front of ``sofa''. But when a common-sense-independent setup occurs, \eg, ``toilet'' exists instead of ``table'', as shown in Fig.~\ref{fig:fig1}~(right), the network tends to give an incorrect answer.
	Based on this observation, we propose the compositional scene manipulation (CSM) strategy, which generates simulated 3D scenes and corresponding questions by modifying the structure of the 3D semantic scene graph.
	Therefore, we further release 127K questions from 7,438 simulated 3D scenes, which provides common-sense-independent solid guidance to the VQA-3D and enhances performance in real-world scenes.
	
	Recent work (ScanQA)~\cite{azuma_2022_CVPR} also proposes a VQA-3D dataset, however, CLEVR3D is different from ScanQA in the following aspects:
	\textbf{1)}~\textbf{Data Generation}: They transform question-answer pairs from the descriptions in the ScanRefer dataset~\cite{chen2020scanrefer} through a T5-base model~\cite{raffel2019exploring}. 
	Our dataset is generated based on 3D semantic scene graphs, where more attributes for objects (\eg, material) and question types (\eg, existence) are provided, as shown in Table~\ref{tab:tab11}.
	\textbf{2)}~\textbf{Data Scale}: 
	Their generation strategy greatly depends on the number of descriptions in the ScanRefer dataset.
	In contrast, we gain theoretically unlimited question-answer pairs through augmenting the scene graph.
	In practice, CLEVR3D contains $4\times$ questions from $11\times$ scenes.
	\textbf{3)}~\textbf{Scenario}: They only focus on VQA-3D in real-world scenes, while we also tackle the problem in simulated common-sense-independent scenes.
	
	In experiments, we baseline several methods on our CLEVR3D dataset.
	Moreover, we propose a framework \textbf{TransVQA3D} based on Transformer~\cite{2017_Vaswani_Attention}, which achieves 3D scene understanding by jointly learning VQA-3D and scene graph generation tasks. Specifically, it introduces explicit relationship supervision to guide the network for better modeling scene context.
	%
	%
	The experiments demonstrate that TransVQA3D achieves superior results on the VQA-3D task against pure language models and current state-of-the-art, whereas the task of CSI-VQA-3D further improves the performance on real-world VQA.

	Our main contributions can be summarized as follows:
	\begin{itemize}
	\setlength{\itemsep}{0pt}
	\setlength{\parsep}{0pt}
	\setlength{\parskip}{.5pt}
	\item  We introduce a large-scale dataset \textbf{{CLEVR3D}} for the task of VQA-3D, where 171K questions from 8,771 real-world 3D scenes are provided.
	\item  We propose a compositional scene manipulation strategy and study the common-sense-independent VQA-3D. 
	\item We baseline several VQA-3D methods, and design {\textbf{TransVQA3D}}, which achieves state-of-the-art results on VQA-3D while boosting the 3D scene graph analysis performance. Moreover, CSI-VQA-3D effectively enhances the performance of VQA-3D.
		\end{itemize}

	\section{Related Work}
	\label{sec:related}
	\subsection{Visual Question Answering on Images}
	VQA has gained significant attention recently.
	This task requires models to answer a text-based question according to the information contained in an image. 
	Several datasets are proposed in this research field, spanning natural images~\cite{2019_Hudson_GQA,2017_Goyal_VQAv2} and synthesis ones~\cite{johnson2017clevr}. 
	Among them, CLEVR~\cite{johnson2017clevr} is a typical diagnostic dataset for image-based VQA.
	It uses 100,100 synthesis images to construct spatial and comparative relationships between different chromatic shapes.
	Inspired but different from CLEVR, our CLEVR3D applies 3D scene graphs to generate more diverse questions.
	Very recently, SimVQA \cite{cascante2022simvqa} use physics simulation platforms to generate synthetic data, but it only contains 35 scenes \cite{gan2020threedworld} and focus on images.
	To answer questions by reference images, early models \cite{2015_Antol_VQA,2016_Kim_MRL,2016_Gao_CBI,2017_Ben-younes_MUTAN,2016_Fukui_MCB,2017_Kim_MLB} design various of features fusion method between vision and language.
	%
	%
	Motivated by the studies of Graph Convolutional Networks (GCN)~\cite{2019_Li_GCN}, several works~\cite{2019_Tang_VCTree,2019_Li_ReGAT,2019_Hudson_NSM} encode visual and linguistic information on graph-based structures.
	Recently, Vision Transformer ~\cite{2017_Vaswani_Attention,2019_Yu_MCAN,2019_Tan_LXMERT,2019_Lu_ViLBERT} has shown its superiority in image-based VQA, and some of these methods use pre-trained BERT~\cite{2019_Devlin_BERT} to boost the performance.
	However, considering the differences in representations and the complexity of 3D scenarios, previous image-based methods cannot be directly applied to 3D scene understanding.
	%
	
			\begin{figure*}[t]
		\begin{centering}
			\center\includegraphics[width=\linewidth]{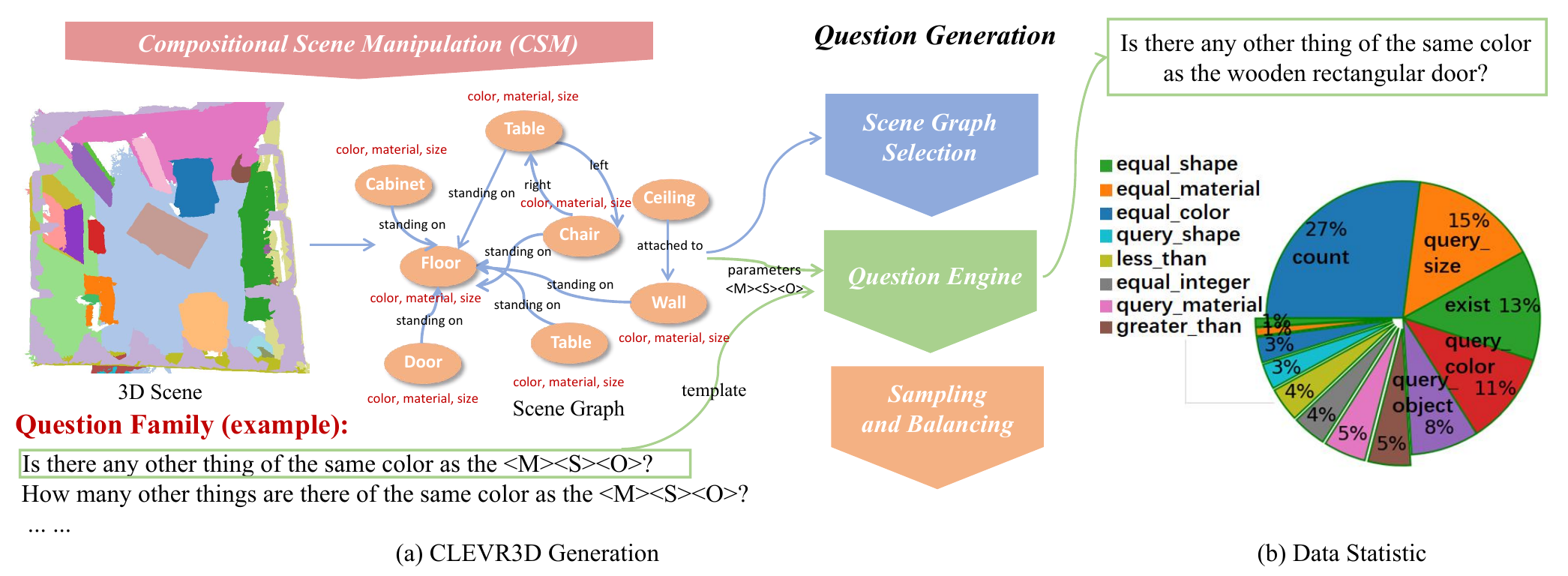}
			\caption{\textbf{Overview of the CLEVR3D.} Part (a) illustrates the data generation process of CLEVR3D-REAL, where the whole process contains three steps: scene graph selection, question engine design, and sampling. Besides, we can further exploit the compositional scene manipulation (CSM) strategy to generate more simulated common-sense-independent 3D scenes and corresponding scene graphs for the CLEVR3D-SIM dataset.
				Part (b) shows the data statistics of question length and proportions. CLEVR3D contains more question types compared with the CLEVR dataset.
			}
			\label{fig:fig2}
		\end{centering}	
	\end{figure*}

	\subsection{Scene Graph Analysis}
	To present object relationships of images in an explicit and structured way, Scene Graph \cite{johnson2015image} is proposed, in which objects are modeled as the nodes, and the edges linking them represent the relationships.
	Since the release of the large-scale 2D scene graph dataset, Visual Genome \cite{2017_Krishna_VG}, a string of scene graph generation methods~\cite{xu2017scene, 2018_Zellers_Motifs, 2019_Wang_MemIMP, 2019_Tang_VCTree, tang2020unbiased, liu2021fully} are substantially fostered, yet following a similar pipeline.
	In practice, a Region Proposal Network (PRN) is utilized to extract the object proposals along with their features.
	Then a fully connected graph refines the node and edge features to infer a scene graph.
	For instance, in the VCTree model~\cite{2019_Tang_VCTree}, the visual features are obtained from Faster-RCNN \cite{2015_REN_Faster_R-CNN}, and a dynamic VCTree is established with a differentiable score matrix.
	Graph embeddings are refined using BiTreeLSTM \cite{tai2015improved}. The context features from graph embeddings are used for scene graph generation and VQA.
	In terms of the 3D scene graph analysis, Johanna \etal~\cite{wald2020learning} is the pioneer.
	They propose the first 3D scene graph benchmark based on the 3RScan dataset~\cite{wald2019rio} and introduce a baseline model exploiting PointNet \cite{qi2017pointnet} to obtain object classes and inter-object relationships.
	Recently, Zhang \etal~\cite{SGGpoint} adopted a graph-based model to enhance the graph analysis.
	Wu \etal~\cite{wu2021scenegraphfusion} proposes a framework for incremental 3D scene graph generation from the RGB-D frames.
	However, these methods mainly focus on the object classes and relationships but ignore object existence and fine-grained objects.

	\subsection{3D Vision and Language}
	3D vision and language understanding is a relatively emerging research field compared to image and language comprehension. 
	Current works focus on using language to confine individual objects, \eg, detecting referred 3D objects~\cite{chen2018text2shape} or distinguishing objects according to language phrases~\cite{achlioptas2019shapeglot}.
	Recently, ScanRefer~\cite{chen2020scanrefer} and ReferIt3D~\cite{achlioptas2020referit3d} introduce a task of localizing objects within a 3D scene given the linguistic descriptions, namely 3D visual grounding.
	Following them, several works are proposed to improve the performance through instance segmentation~\cite{huang2021text,yuan2021instancerefer}, or Transformer~\cite{roh2021languagerefer,yang2021sat,zhao20213dvg}.
	3D dense captioning has been proposed very lately in Scan2Cap~\cite{chen2021scan2cap}.
	It focuses on decomposing 3D scenes and describing the chromatic and spatial information of the objects~\cite{yuan2022x}.
        After that, some works \cite{chen2021d3net, cai20223djcg} try to jointly learn 3D visual grounding and captioning, since there are some intrinsic connections between the two tasks.
        For example, they both require object proposals and cross-modal fusion. Thus, they may benefit each other by training simultaneously. 
	%
	Very recently, \cite{azuma_2022_CVPR} propose ScanQA dataset for the VQA-3D on 3D real-world scenes.
	However, their question-answer pairs are transformed from ScanRefer~\cite{chen2020scanrefer}, and thus they cannot generate sufficient data with variant objects' attributes.
	In contrast, CLEVR3D is generated from 3D scene graphs, where more attributes and question types are included.

	\section{CLEVR3D}
	\label{sec:data}
	In this paper, we propose the CLEVR3D dataset, which contains two sub-datasets, \ie, (a)~\textbf{CLEVR3D-REAL} and (b)~\textbf{CLEVR3D-SIM}, and are used to examine VQA-3D for real-scene and CSI-VQA-3D.  
	In Section~\ref{sec:31}, we introduce the data generation process, including the Selection of scene graphs, the design of the question engine to generate diverse questions via templates, and the way to eliminate data bias.
	In Section~\ref{sec:32}, we provide the data statistics of CLEVR3D and make comparisons with the previous VQA dataset: CLEVR~\cite{johnson2017clevr} and ScanQA~\cite{azuma_2022_CVPR}.

				  \begin{figure*}[t]
		\begin{centering}
			\center\includegraphics[width=\linewidth,height=12cm]{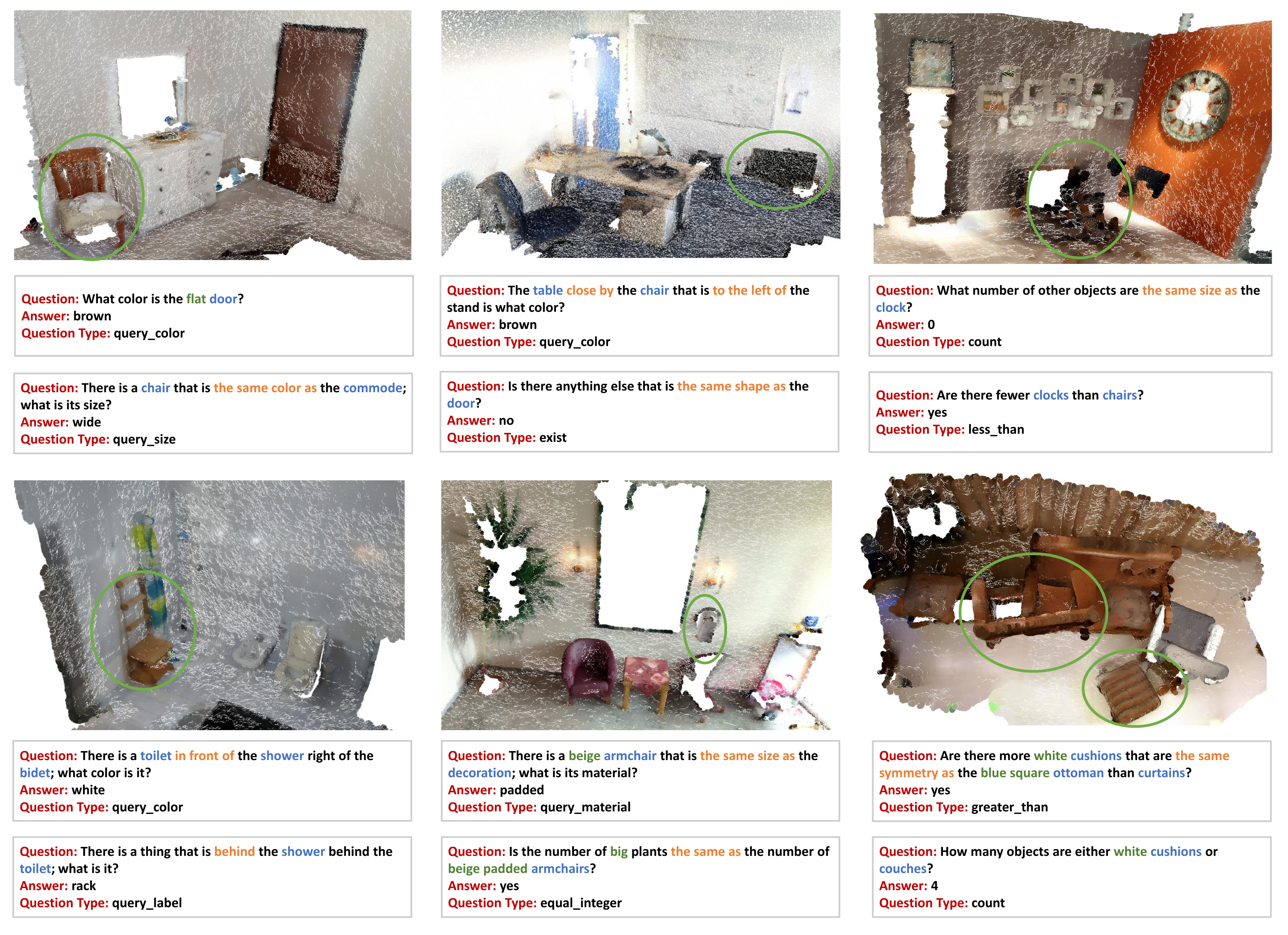}
			\caption{\textbf{Examples in CLEVR3D-SIM.} Green circles indicate the manipulated objects by our proposed compositional scene manipulation strategy.}
			\label{fig:fig6}
		\end{centering}	
	\end{figure*}

	\subsection{Data Generation}
	\label{sec:31}
	\noindent\textbf{Scene Graph Selection.} 
	The foundation of CLEVR3D dataset is the 3D Semantic Scene Graph (3DSSG) annotations~\cite{wald2020learning} in 3RScan dataset~\cite{wald2019rio}.
	The 3RScan dataset was initially proposed for the 3D object instance re-localization and then scene-graphs were introduced in 3DSSG to build a benchmark.
	Following 3DSSG, there are 1,333  scenes that have scene graph annotations.
	%
	A semantic scene graph $G$ in 3DSSG, is a set of tuples $(\mathcal{V}, \mathcal{E})$ between nodes $\mathcal{V}$ and relation edges $\mathcal{E}$.
	Each node denotes an object in the 3D scene, \eg, \textit{table}, \textit{chair} and \textit{window}, and it is linked to a bounding box indicating its position and size.
	Each object is associated with its attributes, \ie, color, shape, and material, and objects are connected by relation edges, representing spatial relations and comparatives.
	Following~\cite{wald2020learning}, we keep 160 object classes and 26 inter-object relationships to alleviate the class imbalance issues, rather than the original 534 object classes and 40 relationships.

	\noindent\textbf{Question Engine.} 
	The question engine is responsible for generating diverse questions.
	We follow the process of traditional VQA dataset generation~\cite{johnson2017clevr}, exploiting \textit{question family} to generate questions.
	Specifically, a question family includes functional programs and text templates for constructing questions, and one can change its parameters to express diverse questions. 
	For instance, the question \textit{``How many white wooden table are there?''} can be formed by the template \textit{``How many $<$C$>$ $<$M$>$ $<$O$>$ are there?''}, exploiting the parameters $<$C$>$, $<$M$>$ and $<$O$>$ (correspond to the values \textit{white}, \textit{wooden} and \textit{table}) to indicate the attributes ``color'', ``material'' and ``object'', respectively.
	In practice, there are five parameters, namely $<$C$>$ (color), $<$M$>$ (material), $<$S$>$ (shape), $<$R$>$ (relationship) and $<$O$>$ (object).
	We designed 90 question families in CLEVR3D, and each has an average of four parameters.
	As Fig.~\ref{fig:fig2}~(a), by applying different families, diverse questions are generated.

		\begin{figure*}[t]
	\begin{centering}
		\center\includegraphics[width=0.99\linewidth]{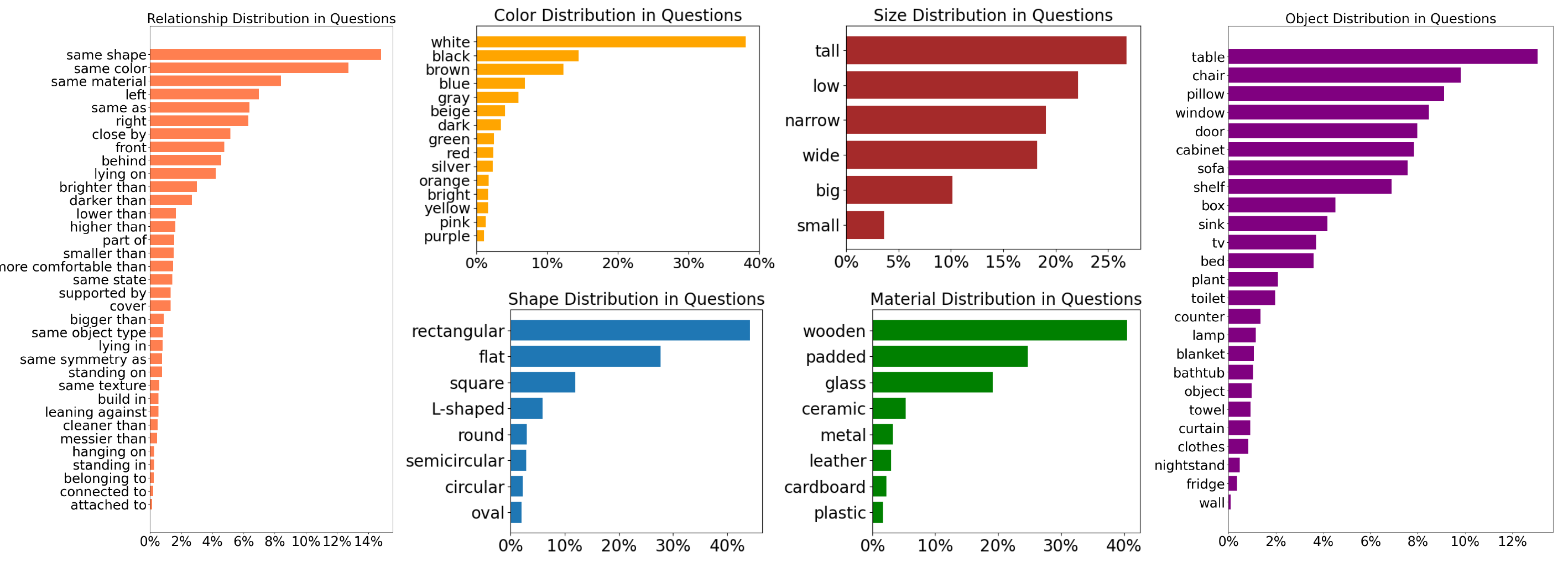}
		\caption{\textbf{The distribution of questions.} We demonstrate the distribution of relationship, color, material, size, shape and object in questions.}
		\label{fig:subfig2}
	\end{centering}	
\end{figure*}

	\noindent\textbf{Question Sampling and Balance.} 
	We randomly sample combinations of values during the question generation and reject those leading to ill-posed or degenerated questions.
	For example, we eliminate questions of the form \textit{``What color is the $<$O$_1>$ to the $<$R$>$ of the $<$O$_2>$''} with $<$O$_1>$=``chair'' and $<$O$_2>$=``table'' if the scenes do not contain ``chair'' or ``table''.
	To further reduce the ambiguity, we do not consider the meaningless objects, \eg, ``floor'' and ``other objects'', in the question since the relationship with them is almost singular (every object ``standing on'' the floor).
	%
	%
	Besides, we eliminate answers that appear to be less than one hundred.

	\begin{figure}[t]
	\begin{centering}
		\center\includegraphics[width=0.8\linewidth]{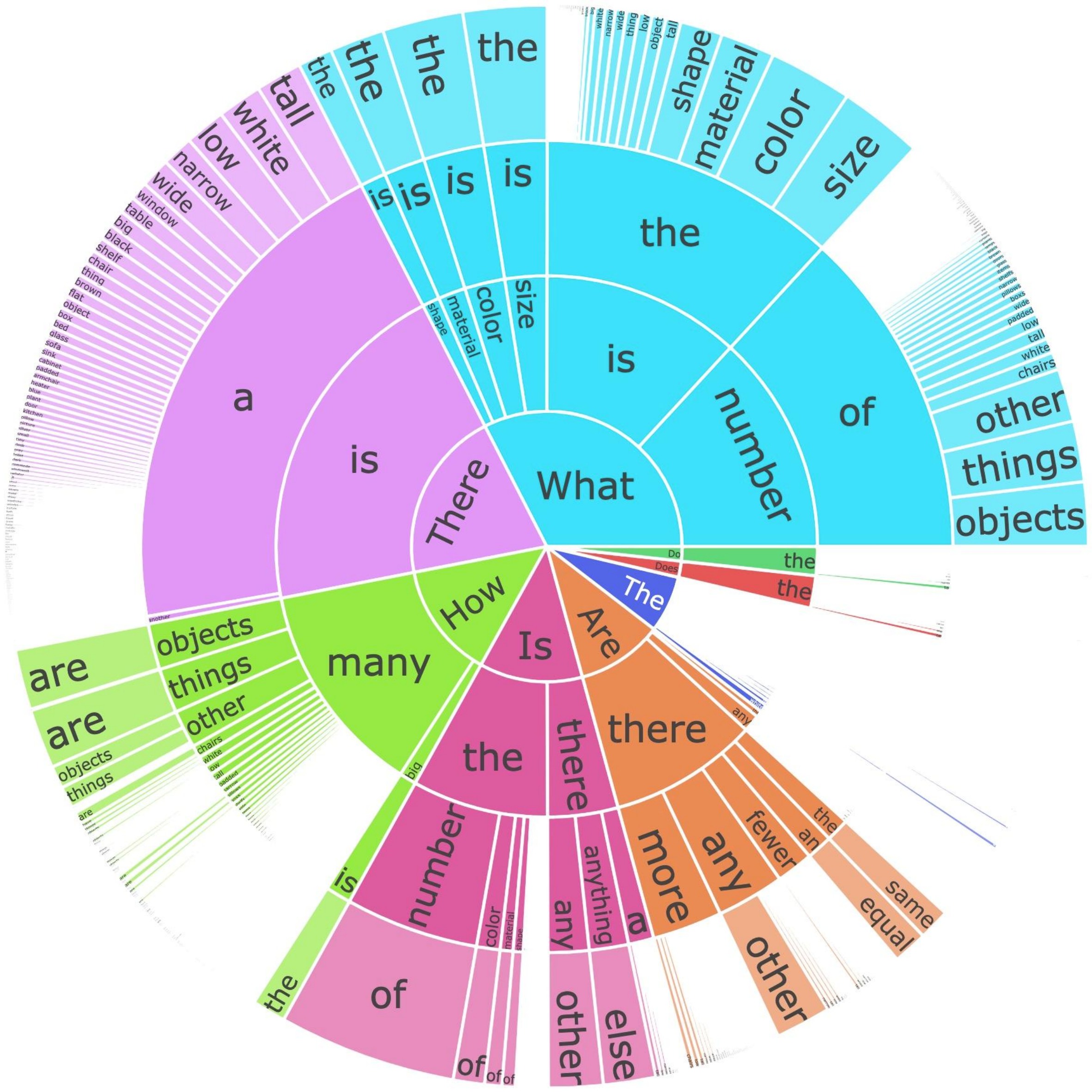}
		\caption{Distribution of the question types by the beginning of the question writing in CLEVR3D-REAL.}
		\label{fig:data}
	\end{centering}	
	\end{figure}

	\noindent\textbf{Compositional Scene Manipulation.}
	Due to some objects commonly co-occurrence in real-world 3D scenes (\eg, table and chair), such prior knowledge could be confounding factors that mislead the VQA model to learn spurious associations.
	In such a situation, the network will fail to answer those `uncommon' questions, \eg, a `toilet' exists behind the `table'.
	Hence, a more effective VQA model is desired to not only handle the common questions but also deal with the more challenging common-sense-independent ones. 
	To generate the dataset of CLEVR3D-SIM for the common-sense-independent VQA-3D (CSI-VQA-3D), a compositional scene manipulation (CSM) strategy is introduced during the data generation process.
	Specifically, we firstly offline preserve all objects with their node attributes in 3DSSG, obtaining an \textit{Object Pool}.
	In the following, we randomly replace each node in the scene graphs with an object in the object pool.
	To ensure the simulated scene is more realistic, we normalize the orientation of each object and place it with the original head orientation.
	Furthermore, we calculate the bounding boxes of both the replaced object and the original one, ensuring that the ratio of volumes is within the range of [1,1.5].
	Finally, for each 3D scene, we use the above CSM operation to generate $10\times$ simulated scenes and manually discard those scenes with unreasonable occlusion or overlap.
	The remaining scenes are used by the question engine.
	Selected examples in CLEVR3D-SIM are illustrated in Fig.~\ref{fig:fig6}.

	%
	%
	%
	%
	
	\begin{figure*}[t]
		\begin{centering}
			\center\includegraphics[width=\linewidth]{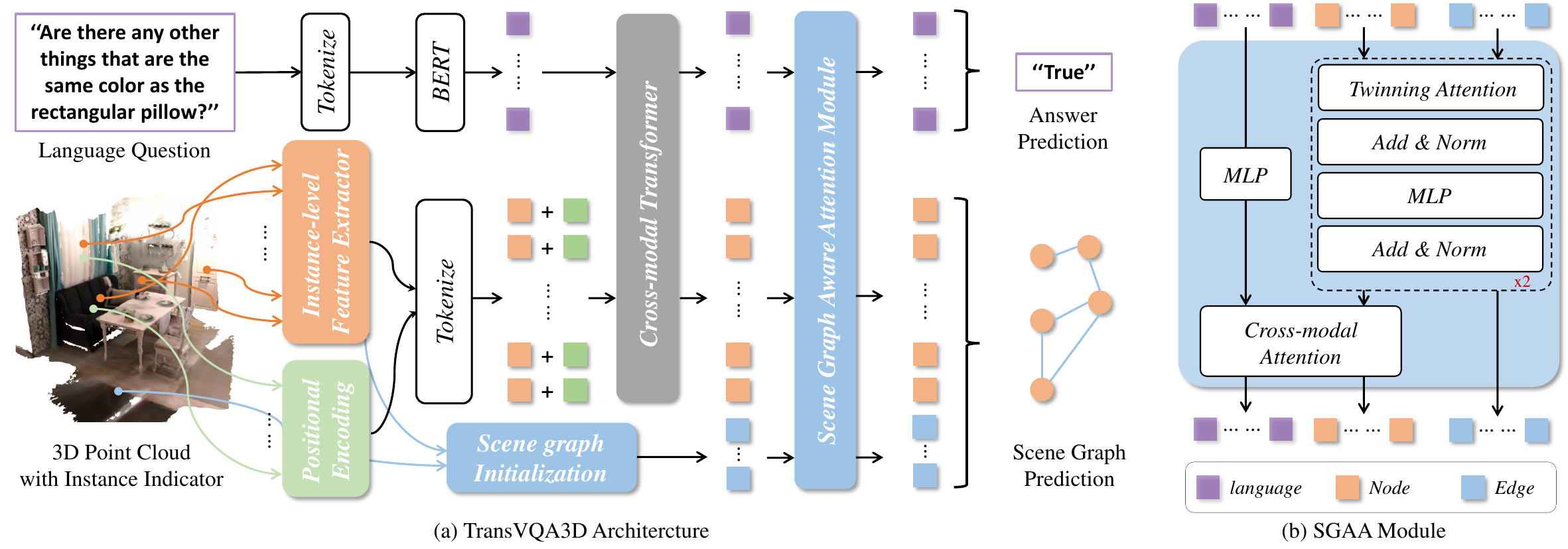}
			\caption{\textbf{Overview of the architecture of TransVQA3D.} (a) The pipeline of TransVQA3D, where TransVQA3D first uses a cross-modal Transformer to fuse the features of language and object. Then, we apply scene graph initialization and take additional edges of the scene graph to conduct scene graph aware attention (SGAA). (b) The inner structure of SGAA module.}
			\label{fig:fig3}
		\end{centering}	
	\end{figure*}

	\subsection{Data Statistics}
	\label{sec:32}
	\noindent\textbf{Data Split.} Our CLEVR3D contains 171,174 questions from 8,771 scenes.
	Specifically, there are two sub-datasets, namely CLEVR3D-REAL and CLEVR3D-SIM.
	For CLEVR3D-REAL, the training and test sets include 38,806 and 5,765 questions from 1,176 and 157 scenes, respectively.
	For CLEVR3D-SIM, there are 109,351 and 17,252 questions from 6,510 and 928 simulated scenes for train and test.

	\noindent\textbf{Question Types.} We categorize questions into different question types, defined by the outermost function in the question’s program, as shown in Fig.~\ref{fig:fig2}~(b).
	Specifically, Fig.~\ref{fig:fig2}~(b) illustrates question types (\eg,``query-color'' and ``count'') and the ratio of questions within each type.
	Moreover, we show the distribution of the question types by the beginning of the question writting in Fig.~\ref{fig:data}.
	
	\noindent\textbf{Attribute Distribution.} We demonstrate the distribution of each attribute (\eg, relationship, color and \textit{etc}.) in Fig.~\ref{fig:subfig2}.
	
%
%
%
%
%
%
%
%
%
	
	\noindent\textbf{Comparison with Existed Dataset.} 
	To further demonstrate the data amount of our CLEVR3D, we compare it with two existed datasets, \ie, CLEVR~\cite{johnson2017clevr} and ScanQA~\cite{azuma_2022_CVPR}.
	Table~\ref{tab:tab1} lists the comparison between CLEVR3D and the most typical VQA-2D dataset, \ie, CLEVR~\cite{johnson2017clevr}, which is similar to our dataset since images in others contain humans and relationships of behaviors.
	As shown in the table, CLEVR3D contains a wider range of categories for objects' attributes, classes, and interrelationships compared with \cite{johnson2017clevr}.
	Furthermore, the comparison with ScanQA~\cite{azuma_2022_CVPR} is shown in Table~\ref{tab:tab11}, where our dataset contains $4\times$ questions from $11\times$ scenes.
	%

	\begin{table}[t]
	\centering
	\caption{Statistic of CLEVR and ours. We demonstrate the total number of categories in attributes, objects, and relationships.}
	\begin{tabular}{l|c|c}
		\toprule
		Dataset & CLEVR~\cite{johnson2017clevr} & CLEVR3D \\\hline
		Number of size  & {2}  & \textbf{6} \\
		Number of color  & {8}  & \textbf{15} \\
		Number of material  & {2}  & \textbf{8} \\
		Number of shape  & {3}  & \textbf{8} \\
		Number of object & {-}  & \textbf{160} \\
		Number of relation  & {4}  & \textbf{26} \\
		\toprule
	\end{tabular}%
	
	\label{tab:tab1}
\end{table}

\begin{table}[t]
	\centering
	\caption{Statistical comparison with CLEVR and ScanQA. }
	\begin{tabular}{l|ccc}
		\toprule
		Dataset & CLEVR~\cite{johnson2017clevr} & ScanQA~\cite{azuma_2022_CVPR}  &CLEVR3D \\\hline
		descriptor & 100,100 images & 800 scans  & 8,771  scans    \\
		\#question & 853,000  &41,363 & 171,174 \\ \hline
		avg \#question& {9} & {52} & {20} \\
		avg length & {18} & {8} & {12}  \\
		question types & 13 & 6 & 13 \\
		\toprule
	\end{tabular}%
	
	\label{tab:tab11}
\end{table}

	\section{Method}
	\label{sec:method}
	In this paper, we propose a strong baseline for the VQA-3D task built upon Transformer~\cite{2017_Vaswani_Attention}.
	Fig.~\ref{fig:fig3}~(a) illustrates the architecture of our proposed TransVQA3D, which contains three modules, \ie, object embedding, cross-modal transformer, and scene graph aware attention (SGAA) module.
	Section~\ref{sec:41} introduces the input embedding, including feature embedding and positional encoding.
	The details of cross-modal transformer and SGAA is addressed in the Section~\ref{sec:42} and Section~\ref{sec:43}.

	\subsection{Input Embedding}
	\label{sec:41}
	The input to a VQA-3D model is a 3D scene $\mathcal{P} \in \mathbb{R}^{N\times D_{\text{in}}}$ (in the form of point clouds with $N$ points and a $D_{\text{in}}$-dimensional feature on each point) and a natural language question with $l$ words.
	To facilitate the setup, we only use the coordinate (XYZ) of each point and its color values (RGB) as a 6D input vector while ignoring additional hints such as normal vectors and projected 2D features~\cite{yang2021sat}.
	
	\noindent\textbf{Language Embedding.} 
	The input query description is first tokenized into words and further mapped into a sequence of vectors $\mathcal{W} \in \mathbb{R}^{l\times 768}$ via a pre-trained BERT~\cite{2019_Devlin_BERT}, where $l$ is the sequence length.
	To reduce the computational burden during training, a fully-connected layer maps the feature dimension to $d_{\text{enc}}$, following a dropout layer of ratio 0.1.
	
	\noindent\textbf{Object Embedding.} 
	Following previous studies~\cite{wald2020learning}, we assume that there are $m$ 3D object instances in the scene, and we use a class-agnostic point-to-instance indicator $\mathcal{M} \in \{1, ...,m\}^N$ to generate instance point clouds $\mathcal{P}^{obj}=\{\mathcal{P}^{obj}_i\}^m$ for each object $i$:
	\begin{equation}
	\mathcal{P}_i^{\text{obj}} = \{\mathcal{P}_n | \mathcal{M}_n=i; ~{n=1,...,N}\}.
	\label{eq1}
	\end{equation}
	%
	%
	After that, a share-weighted feature extractor~\cite{qi2017pointnet} $\mathcal{F}^{\text{obj}}(\cdot)$ followed with a symmetric pooling function is performed on each instance point cloud to obtain instance feature, \eg, $\mathcal{X}_i^{\text{pc}} = \mathcal{F}(\mathcal{P}^{\text{obj}}_i) \in \mathbb{R}^{d_{\text{enc}}}$.
	To encode spatial information, we further design the object positional encoding $\mathcal{X}^{pos}$ as following:
	\begin{equation}
	\mathcal{X}_i^{\text{pos}} = \mathcal{F}^{\text{pos}}([\mathcal{P}^{\text{center}}_i;~\mathcal{P}^{\text{size}}_i;~\mathcal{P}^{\text{orient}}_i]) \in \mathbb{R}^{d_{\text{enc}}},
	\label{eq2}
	\end{equation}
	where $\mathcal{F}^{\text{pos}}$ is a non-linear transformation function; $\mathcal{P}^{\text{center}}_i \in \mathbb{R}^{3}$, $\mathcal{P}^{\text{size}}_i\in \mathbb{R}^{3}$ and $\mathcal{P}^{\text{orient}}_i\in \mathbb{R}^{1}$ are center, size (height, width, and length), and heading orientation of the object, respectively. $[\cdot; \cdot]$ denotes a concatenation operation.
	\textbf{It should be noted that, since there exists orientation annotation for the part of objects in 3RScan \cite{wald2019rio} dataset, we can reason about view-dependent relationship for these objects} (\eg, `\textit{to the right of}').
	Finally, the embedding of each object can be obtained by merging its feature and positional encoding through
	\begin{equation}
	\mathcal{X}_i = \mathcal{LN}(W_1\mathcal{X}_i^{\text{pc}}) + \mathcal{LN}(W_2\mathcal{X}_i^{\text{pos}}),
	\label{eq3}
	\end{equation}
	where $W_1$, $W_2$ are two learnable projection matrices and $\mathcal{LN}(\cdot)$ indicates layer normalization.
	We form all object embeddings as a set $\mathcal{X} = \{\mathcal{X}_i\}^m \in \mathbb{R}^{m\times d_\text{enc}}$.

	\subsection{Cross-modal Transformer}
	\label{sec:42}
	Inspired by the great success achieved by Transformer~\cite{2017_Vaswani_Attention} in both computer vision and natural language processing, we develop a cross-modal transformer for vision-language feature fusion.
	Specifically, we first concatenate the object embedding with language embedding in the length dimension, obtaining a $m+l$ length token set, where $m$ and $l$ respectively stands for instance number and language length.
	Then, we construct a Transformer by stacking three self-attention layers with four heads.
	After feeding object embedding $\mathcal{X}$ and language embedding $\mathcal{W}$ into the model, the enhanced features $\mathcal{X}'$ and $\mathcal{W}'$ are obtained for performing scene graph aware attention.
	
	\subsection{Scene Graph Aware Attention}
	\label{sec:43}
	The inner structure of the scene graph aware attention (SGAA) module is depicted in Fig.~\ref{fig:fig3}~(b).
	The goal of the SGAA module is to simultaneously obtain the answer and create a scene graph $G = (\mathcal{V}, \mathcal{E})$, where nodes $\mathcal{V}$ and edges $\mathcal{E}$ depict the instances and their inner structural relationships, respectively.
	It contains two twinning attentions and cross-modal attention layers.
	
	\noindent\textbf{Scene Graph Initialization.}
	To obtain a scene graph representation, the node feature is initialized as $\mathcal{X}_{\mathcal{V}} = \mathcal{X}' \in \mathbb{R}^{m\times d_{\text{enc}}}$, which is further propagated to obtain a $m^2$ one-to-one edges $\mathcal{X}_{\mathcal{E}}$. Each edge feature $\mathcal{X}_{\mathcal{E}} \in \mathbb{R}^{m\times m\times (2d_{\text{enc}}+9)}$ contains the features of two nodes ($2d_{\text{enc}}$), and additional $\mathbb{R}^{9}$ features, \ie, two objects' centers and their relative offsets.

	\noindent\textbf{Twinning Attention.}
	To fully utilize the language and scene graph information to mutually enhance the performance of two tasks, we adopt the twinning attention, a graph convolution operator proposed in \cite{SGGpoint}.
	This mechanism takes nodes and edges of a scene graph as the inputs and generates their enhanced representation $\mathcal{X}_{\mathcal{V}}'$ and $\mathcal{X}_{\mathcal{E}_{i, j}}'$. 
	In specific, two steps are introduced for feature updating: edge-to-node and node-to-edge.
	
	\noindent (1) \textit{edge-to-node}: It first updates node features through bidirectional edge attention.
	Given a node pair $(i,j)$, it considers both the relationship from node $i$ to $j$ and the inverse one from $j$ to $i$.
	In particular, for the $i$-th node feature $\mathcal{X}_{\mathcal{V}_{i}}$, it conducts feature aggregation within both row and column of the edges feature (\ie, $\mathcal{X}_{\mathcal{E}_{i,:}}$ and $\mathcal{X}_{\mathcal{E}_{:,i}}$) by 
	\begin{equation}
	\mathcal{R}_i = \Sigma(W_3 \{\mathcal{X}_{\mathcal{E}_{i,j}}\}^m_{j=1}) \odot  \Sigma(W_4 \{\mathcal{X}_{\mathcal{E}_{j, i}}\}^m_{j=1})
	\label{eq4}
	\end{equation}
	where $W_3$ and $W_4$ are two linear projection from $\mathbb{R}^{m\times d_{\text{enc}}}$. 
	After independently merging row and column features into single vectors through a pooling function $\Sigma(\cdot)$, it applies Hadamard product $\odot$ to obtain the relation representation $\mathcal{R} \in \mathbb{R}^{m\times d_{\text{enc}}}$.
	Then, the node feature is updated by 
	\begin{equation}
	\mathcal{X}_{\mathcal{V}}' = f(\mathcal{X}_{\mathcal{V}} \odot \sigma(\mathcal{R})).
	\label{eq5}
	\end{equation}
	In Eqn.~\eqref{eq5}, a Sigmoid function $\sigma(\cdot)$ is first conducted on edge relation and obtained edge-driven interactive scores.
	After applying the product between node and relation, a nonlinear activation function $f$ is further adopted.
	\noindent (2) \textit{node-to-edge}: It updates the edge feature through the nonlinear transformation $f$ of the concatenation of newly generated node feature:
	\begin{equation}
	\mathcal{X}_{\mathcal{E}_{i, j}}' = f(W_5 [\mathcal{X}_{\mathcal{V}_i}';~\mathcal{X}_{\mathcal{V}_j}']).
	\label{eq6}
	\end{equation}
	
		\begin{table*}[t]
		\centering
		\caption{Accuracy per question type of the different VQA-3D methods on the CLEVR3D-REAL.}
		\resizebox{\textwidth}{!}{
			\begin{tabular}{l|cccccc|c}
				\toprule
				Method & {Existence} & {Counting} & {Compare Integer} & {Query Attr.} &{Query Object} & {Compare Attr.} & {Overall} \\
				\textit{Number} & \textit{474} & \textit{1,386} & \textit{602} & \textit{2,339} & \textit{355} & \textit{609} & \textit{5,765} \\\hline\hline
				L+LSTM & 86.5 & 26.8 & 56.6 & 39.4 & 19.2 & 55.5 & 42.5 \\
				L+Transformer & 86.3 & 25.4 & 59.3 & 38.1 & 13.2 & 56.5 & 41.6 \\\hline
				ReferIt3D~\cite{achlioptas2020referit3d} & 85.4     & 26.6     & 57.5     & 38.3     & 11.3     & \textbf{58.5}  & 41.8 \\
				BEV+MCAN~\cite{yu2020deep} & 85.7    & 27.3     & 60.0    & 39.5     & 14.4     & 53.7 & 42.5 \\
				ScanQA~\cite{azuma_2022_CVPR} & 84.8    & 26.3     & {55.6}    & \textbf{41.6}    & 13.2     & 54.7 & 42.6 \\
				L+V+Transformer & 87.6 & 26.4 & 55.5 & 39.8 & 21.4 & 58.3 & 43.0 \\
				TransVQA3D & \textbf{88.2}    & 30.7     & 63.1     & 37.3     & 21.7     & 57.1 & 43.7 \\
				TransVQA3D+BEV & 86.1    & 29.4     & 62.1     & 40.8     & 17.5     & 56.8 & 44.3 \\
				{w/} CLEVR3D-SIM & {87.1}     & \textbf{32.5}     & \textbf{65.9}     & {39.7}     & \textbf{22.8}     & 57.5 & \textbf{45.5} \\\hline
				Human &  90.3     & 64.1      & 71.8      & 75.6      & 60.7      & 73.9      & 72.6      \\
				\toprule
			\end{tabular}%
		}
		\label{tab:tab2}%
	\end{table*}%
	
	\begin{table*}[t]
		\centering
		\caption{Accuracy per question type of the different VQA-3D methods on the CLEVR3D-SIM.}
		\resizebox{\textwidth}{!}{
			\begin{tabular}{l|cccccc|c}
				\toprule
				Method & {Existence} & {Counting} & {Compare Integer} & {Query Attr.} &{Query Object} & {Compare Attr.} & {Overall} \\
				\textit{Number} & \textit{2,316} & \textit{4,271} & \textit{2,363} & \textit{5,681} & \textit{1,472} & \textit{1,149} & \textit{17,252} \\\hline\hline
				L+LSTM & \textbf{87.8}     & 50.1     & 63.0     & 31.8    & 17.4     & 53.9 &  48.4 \\
				L+V+Transformer & 87.4 & \textbf{55.4} & \textbf{69.1} & 30.8 & \textbf{21.0} & \textbf{54.8} & 50.5 \\
				TransVQA3D & 87.6    & 54.5     & 68.2     & \textbf{34.5}     & 19.3     & 53.9 & \textbf{51.2} \\
				\toprule
			\end{tabular}%
		}
		\label{tab:tab22}%
	\end{table*}%

	\noindent\textbf{Cross-modal Attention.}
	To leverage the scene graph aware representation for the answer prediction, a cross-modal attention mechanism is applied in the SGAA module. 
	For each language feature $\mathcal{W}_{i}'$, it aggregates the node features through
	
	\begin{align}
	\hat{\mathcal{W}}_{i} = \Sigma(\{\sigma(W_3[\mathcal{W}_{i}';~\mathcal{X}_{\mathcal{V}_j}']) \odot \mathcal{X}_{\mathcal{V}_j}', ~~ \forall~\mathcal{X}_{\mathcal{V}_j}' \in \mathcal{X}_{\mathcal{V}}'\}),
	\label{eq7}
	\end{align}
	
	where the language feature $\mathcal{W}_{i}'$ is firstly concatenated with node feature $\mathcal{X}_{\mathcal{V}_j}'$ and then obtain normalized weights through Sigmoid function $\sigma(\cdot)$.
	After that, it aggregates the node features through weighted average and pooling $\Sigma(\cdot)$.
	
	\subsection{Prediction and Objectives}
	As shown in Fig.~\ref{fig:fig3}~(a), TransVQA3D generates three kinds of outputs, including enhanced features for language, nodes, and edges, respectively.
	To facilitate the notation, we denote them as $\hat{\mathcal{W}}$, $\hat{\mathcal{X}_{\mathcal{V}}}$ and $\hat{\mathcal{X}_{\mathcal{E}}}$, respectively.
	$\hat{\mathcal{W}}$ is used in the answer prediction by passing through several fully-connected layers, obtaining scores toward all candidate answers in the answer pool.
	$\hat{\mathcal{X}_{\mathcal{V}}}$ and $\hat{\mathcal{X}_{\mathcal{E}}}$ are independently fed into two classifiers, and the probabilities of object classes (nodes) and object-to-object relationships (edges) can be gained.
	The cross-entropy loss is used as the criterion for answer prediction and node classification ($\mathcal{L}_{vqa}$ and $\mathcal{L}_{node}$).
	Since the edges of a scene graph generally have multiple labels, we use multi-label binary cross-entropy ($\mathcal{L}_{edge}$) on this branch.
	The final loss is a linear combination of these three losses.

	\begin{figure}[t]
		\centering
            \includegraphics[width=0.95\linewidth]{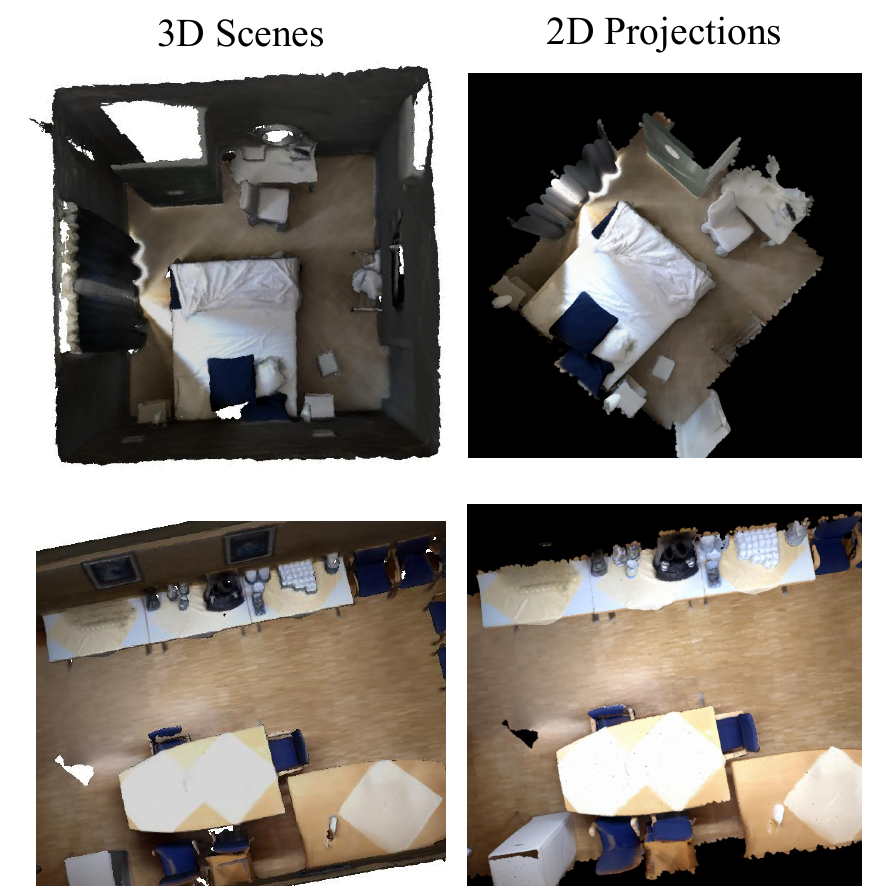}
		\caption{Demonstration of 3D-to-2D projections. To adopt previous 2D-based methods in 3D scenarios, we manually capture 2D images from the 3D meshes. Specifically, we take the bird's eye view (BEV) as it can represent the whole scene with one shot. To eliminate the occlusion problem, we ignore the ceiling and wall that are unrelated to the questions.}
		\label{fig:2d3d}
	\end{figure}
	
	\section{Experiments}
	In this section, we benchmark the VQA-3D while analyzing scene graph understanding and network architecture. 
	%
	%
	\label{sec:exp}
	\subsection{Implementation Details}
	In the implementation, we employ the pre-trained BERT~\cite{2019_Devlin_BERT} to generate the initial language features.
	For the object features, we exploit PointNet++~\cite{qi2017pointnet++} as the feature extractor.
	Each object contains 4,000 points through random sampling, and they are randomly rotated along the z-axis for augmentation.
	Cross-modal Transformer consists of three four-head attention layers, and the output of each layer preserves the identical 256 channels.
	The SGAA module includes two twinning attentions and cross-modal attention, where their hidden dimensions are the same as that of the cross-modal Transformer.
	We train the network for 20 epochs using the Adam optimizer with a batch size of 32 (about 6 hours). 
	The learning rate of the network is initialized as 0.0001 with a decay of 0.9 for every ten epochs. 
	According to the experiment results, the weights of three losses do not significantly affect the performance, and we eventually set them identical.
	 All experiments are implemented on PyTorch with a single RTX2080 GPU.

	\subsection{VQA-3D}
	\noindent\textbf{Dataset.}
	Our proposed CLEVR3D dataset is used to evaluate the performance of different methods for VQA-3D.
	The training and test sets include 49,650 and 10,455 questions in 996 and 133 scenes, respectively.
	We train all methods on the training set and evaluate their performance on the test set.
	
	\noindent\textbf{Baselines.}
	To evaluate the VQA-3D task, we introduce several baselines, including
	\textbf{\textcolor[rgb]{0,0.3,0.7}{pure language models}}, \textbf{\textcolor[rgb]{0,0.4,0}{vision-language cross-modal models}} and  \textbf{\textcolor[rgb]{1,0.5,0.3}{humans results}}.
	Table~\ref{tab:tab2} demonstrates the comparison results among these methods.
	
	\noindent$\blacklozenge$~\textit{\textbf{\textcolor[rgb]{0,0.3,0.7}{L+LSTM}}}: We feed language embedding into a two-layer LSTM. The hidden state of the last layer is passed to a classifier to predict the question answer.
	
	\noindent$\blacklozenge$~\textit{\textbf{\textcolor[rgb]{0,0.3,0.7}{L+Transformer}}}: We remain the language-related components in TransVQA3D, including language embedding and Transformer layers, but discard the 3D point cloud input. The predictions solely rely on language features.
	
	\noindent$\blacklozenge$~\textit{\textbf{\textcolor[rgb]{0,0.4,0}{BEV+MCAN}}}: To adopt previous 2D-based methods~\cite{yu2020deep} in 3D scenarios, we manually capture 2D images from the 3D meshes. Specifically, we take the bird's eye view (BEV) as it can represent the whole scene with one shot, as shown in Fig.~\ref{fig:2d3d}. To eliminate the occlusion problem, we ignore the ceiling and wall that are unrelated to the questions. We fine-tune and test the MCAN model \cite{2019_Yu_MCAN} on these images since it is representative of 2D VQA models.
	
	\noindent$\blacklozenge$~\textit{\textbf{\textcolor[rgb]{0,0.4,0}{ReferIt3D}}}: This model \cite{achlioptas2020referit3d} is initially designed for 3D visual grounding. It firstly extracts object features through PointNet++~\cite{qi2017pointnet++}, then concatenates the global language features with each object representation, and exploits DGCNN~\cite{phan2018dgcnn} to refine the instance features. We discard its scoring step and apply max pooling to obtain a global feature, then pass it to the classifier for question answering.
		
	\begin{table*}[t]
		
		\centering
		\caption{Scene graph analysis on 3DSSG dataset. Top-K Recall is utilized as the evaluation metric. We compare different methods on node classification, predicate classification, and relationship triplet.}
		\begin{tabular}{l|cc|cc|cc}
			\toprule
			\multirow{2}[0]{*}{Method} & \multicolumn{2}{c|}{Object Classification} & \multicolumn{2}{c|}{Predicate Classification} & \multicolumn{2}{c}{Relationships Triplet} \\
			& R@5 & R@10 & R@3 & R@5 & R@50 & R@100\\
			\hline
			SGPN~\cite{wald2020learning} & 66.2  & 77.5 & 62.5 & 88.8 & 39.2 & 45.5 \\
			3DSSG~\cite{wald2020learning} & 68.0   & 78.1 & 89.0 & 93.4 & 40.9 & 66.8 \\
			SceneGraphFusion \cite{wu2021scenegraphfusion} & 70.0 & 80.0 & 97.0 & 99.0 & 85.0 & 87.0 \\
			\hline
			TransVQA3D {w/o} VQA-3D & 69.2   & 79.1 & 97.3 & 99.5 & 86.8 & 88.5 \\
			TransVQA3D &   \textbf{71.8}    &   \textbf{80.6}    &   \textbf{98.0}    &    \textbf{99.8}   &    \textbf{87.1}   &  \textbf{89.8} \\
			\toprule
		\end{tabular}
		\label{tab:tab3}
	\end{table*}
	
	\begin{figure*}[t]
		\begin{centering}
			\centering\includegraphics[width=\linewidth]{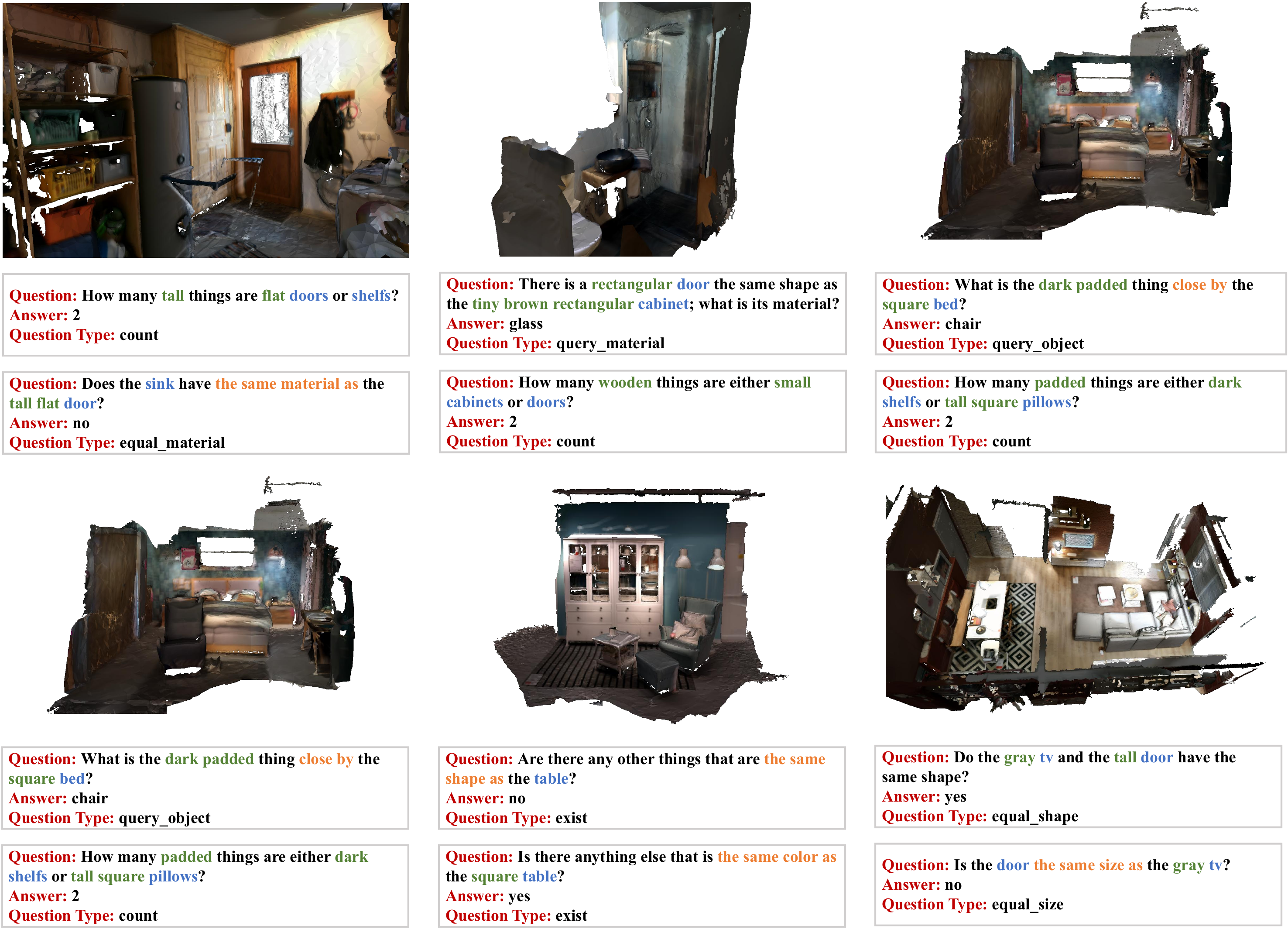}
			\caption{{Visualization of accurate predictions of TransVQA3D on CLEVR3D-REAL.}}
			\label{fig:fig4}
		\end{centering}	
	\end{figure*}

	\begin{table*}[t]
		\centering
		\small
		\caption{Ablation studies for multitask learning of VQA-3D and scene graph analysis. The upper, middle, and lower parts demonstrate the effectiveness of exploiting multitask, different designs for SGAA module and Cross-modal Transformer, respectively.}
		\resizebox{\textwidth}{!}{
			\begin{tabular}{c|c|cc|cc|cc|cc}
				\toprule
				\multirow{2}{*}{Model} & \multirow{2}{*}{Task} & \multicolumn{2}{c|}{CMT} & \multicolumn{2}{c|}{SGAA} & \multicolumn{2}{c|}{SG Analysis} & \multicolumn{2}{c}{VQA-3D} \\
				&    & NumLayer & PE & TA & CMA & Node (\textit{R@5}) & Edge (\textit{R@5}) & Overall & ClassAvg \\
				\hline
				\hline
				A     & VQA-3D   & 3     & \cmark &     -  &    -  &    -   &   -    &    39.7   & 40.1 \\
				B     & VQA-3D+Node   & 3     & \cmark &     -  &    -  &    68.3   &   -    &    42.7   & 45.6 \\
				C     & Node+Edge   & 3     & \cmark &    \cmark  &     \cmark  &    66.7   &    94.7  &    -  &- \\
				D     & VQA-3D+Node+Edge & 3     & \cmark & \cmark & \cmark &   \textbf{71.8}   &   \textbf{99.8}    &   \textbf{45.5}    &  \textbf{50.9} \\
				\hline
				E     & VQA-3D+Node+Edge & 3     & \cmark & \xmark & \cmark &   67.9    &    65.3   &    42.5   & 44.3 \\
				F     & VQA-3D+Node+Edge & 3     & \cmark & \cmark & \xmark &   71.4    &    99.5   &    43.7   & 47.5 \\
				\hline
				G     & VQA-3D+Node+Edge & 4     & \cmark & \cmark & \cmark &    70.2  &    97.6   &   44.5   & 47.4 \\
				H     & VQA-3D+Node+Edge & 2     & \cmark & \cmark & \cmark &     68.5  &    95.8   &   43.2   & 45.1 \\
				I     & VQA-3D+Node+Edge   & 3     & \xmark &     \cmark  &     \cmark  &    67.1   &    94.5   &    42.1   & 44.9 \\
				\bottomrule
			\end{tabular}
		}
		\label{tab:tab4}
	\end{table*}

	\noindent$\blacklozenge$~\textit{\textbf{\textcolor[rgb]{0,0.4,0}{ScanQA}}}: This approach~\cite{azuma_2022_CVPR} is currently state-of-the-art for the VQA-3D. It joint learns visual question answering and object localization at the same time. We use its encoding layers and fusion layers to obtain answer predictions but discard its localization part.

	\noindent$\blacklozenge$~\textit{\textbf{\textcolor[rgb]{0,0.4,0}{L+V+Transformer}}}: We adopt language and visual components of TransVQA3D while ignoring the scene graph aware attention (SGAA) module.
	
	\noindent$\blacklozenge$~\textit{\textbf{\textcolor[rgb]{0,0.4,0}{TransVQA3D}}}: The full architecture described in Section~\ref{sec:method}. 
	
	\noindent$\blacklozenge$~\textit{\textbf{\textcolor[rgb]{0,0.4,0}{TransVQA3D+BEV}}}: We further improve the performance of our method by utilizing additional 2D features from BEV images which is the same as Fig.~\ref{fig:2d3d}.
	Specifically, we adopt 2D CNN to extract the feature from each BEV image, then use global max pooling to obtain the global scene features. We concatenate it with object embeddings in length dimension and feed them into the transformer layers, which can enhance the representation learning for 3D objects.
	
	\noindent$\blacklozenge$~\textit{\textbf{\textcolor[rgb]{0,0.4,0}{TransVQA3D with CLEVR3D-SIM}}}: We pre-trained the above TransVQA3D+BEV model on the CLEVR3D-SIM dataset, and fine-tune on the CLEVR3D-REAL.
	
	\noindent$\blacklozenge$~\textit{\textbf{\textcolor[rgb]{1,0.5,0.3}{Human}}}: Besides these baseline models, we also report the responses of 4 human subjects for 1,000 randomly selected questions from the test set, selecting the answer with the majority of votes among four voters for each question.

	\noindent\textbf{Comparison.}
	We adopt \textit{Accuracy} as the evaluation metric and present the results of different question types on the CLEVR3D-REAL dataset are shown in Table~\ref{tab:tab2}.
	TransVQA3D achieves the best result almost in all question categories, significantly surpassing the pure language model.
	It should be noted that utilizing the feature fusion in ReferIt3D~\cite{achlioptas2020referit3d} cannot effectively improve the VQA-3D performance.
	Without the well-designed architecture and scene graph aware feature enhancement, the performance of ScanQA~\cite{azuma_2022_CVPR} is significantly lower than ours.
	Notably, there still exists a gap over the human responses, especially for ``\textit{count}'' and ``\textit{query object}''.
	Some demos of TransVQA3D for VQA-3D are visualized in Fig.~\ref{fig:fig4}.
	Moreover, CLEVR3D-SIM can effectively boost the performance of the model in real-world VQA-3D, as shown in Table~\ref{tab:tab2}. We also illustrate the results directly training and testing on CLEVR3D-SIM in Table~\ref{tab:tab22}.
	Note that the results on CLEVR3D-SIM  is higher than that on CLEVR3D-REAL since the data amount on the former is much larger.
	
	\subsection{Scene Graph Analysis}
	To further validate the effectiveness of our proposed methods on 3D scene understanding, we evaluate TransVQA3D for scene graph analysis on the 3DSSG dataset~\cite{wald2020learning}.
	We follow the experimental settings of \cite{wald2020learning}, including the train/test split and corresponding semantics.
	We evaluate the model on 160 object and 26 predicate classes with \cite{wald2020learning} in Table~\ref{tab:tab3}. 
	Its vanilla model (\ie, SGPN in the table) uses two PointNets~\cite{qi2017pointnet} to extract the features of the object and relationship, respectively.
	Upon this baseline, it further adopts the GCN to conduct feature aggregation between nodes (\ie, 3DSSG).
	It separately evaluates the predicate (relationship) prediction in isolation from the object classes, where we adopt the Top-k Recall score~\cite{lu2016visual} as the metric.
	Moreover, the performances of the object categories are reported. 
	We further evaluate the most confident \textit{(subject, predicate, object)} triplets against the ground truth in a top-k recall manner. 
	As illustrated in Table~\ref{tab:tab3}, our model effectively improves scene graph analysis in all scene graph-related metrics, especially for relationships triplet.

	 \subsection{Ablation Study}
	 We conduct ablation studies on different designs of our architecture in Table~\ref{tab:tab4}.
	 In the upper part (models A-D), we demonstrate the mutual enhancement between VQA-3D and scene graph reasoning.
	 In the middle part (models E and F), we explore different designs of the SGAA module.
	 In the lower part (models G-I), we present the results of using different Cross-modal Transformer (CMT) designs.
	 Apart from the metric of \textit{Overall Accuracy} (Overall), we also report the results of \textit{Class Average Accuracy} (ClassAvg) to measure the abilities of models crossing different question types.
	
	 \noindent\textbf{Does scene graph analysis help VQA-3D?}
	 We provide the results of only exploiting VQA-3D task in model A, where we do not apply object (node) classification and edge (predicate) prediction.
	 In Table~\ref{tab:tab4}, it is evident that model A gains a low overall accuracy of 39.7 for VQA-3D.
	 When we apply additional supervision of object classification (model B) during the training, the \textit{Overall} boosts from 39.7 to 42.7, which implies the significance of the object classification.
	 Besides, scene graph prediction also help improve the VQA-3D results (model D).
	 There is a noticeable improvement in the absolute accuracy by 2\% when utilizing the SGAA module for edge prediction.
	
	 \noindent\textbf{Does VQA-3D help scene graph analysis?}
	 We show the ablated results by discarding the language branch and answer prediction head during the training.
	 As shown in Table~\ref{tab:tab4} (model C), only conducting scene graph reasoning leads to an obvious degradation on both node and edge prediction (about 5\%).
	
	 \noindent\textbf{How to design the SGAA module?}
	 We independently validate two core components (\ie, twinning Attention (TA) and Cross-modal Attention (CMA) in the SGAA module) in models E and F.
	 In model E, we ablate TA from the SGAA module, leading to a drop on both the VQA-3D and scene graph prediction. 
	 Inversely, the effectiveness of CMA is not prominent since it only affects the results of VQA-3D with about 2\% and the results of scene graph analysis maintain.
	 We ignore the analysis of the number of TA in the SGAA module in this section, since it does not significantly influence the performance within the range of [1,~3].
	
	 \noindent\textbf{How to design Cross-modal Transformer?}
      We test models with different layers in CMT and with/without the position embedding layer.
	 As displayed in the results of model I, positional encoding plays a significant role in the CMT. Without PE, the VQA overall accuracy decreases 3\% and the ClassAvg accuracy  drop 6\%.
	 The number of layers (models G and H) in the CMT does not greatly affect the performances.
	
	\section{Conclusion}
	This paper presents CLEVR3D, a large-scale dataset for enabling the Visual Question Answering task on 3D Point Cloud (VQA-3D).
	We introduce a question engine based on 3D scene graphs to generate diverse reasoning questions concerning object attributes and relationships.
	Moreover, a compositional scene manipulation strategy is leveraged for generating a simulated common-sense-independent dataset CLEVR3D-SIM, avoiding the model to learn spurious or biased associations from limited real-world scenarios. 
	Besides, we propose a baseline model for VQA-3D, namely TransVQA3D, achieving superior performance on this newly proposed dataset and showing that CLEVR3D can significantly help the 3D scene understanding.
	We believe this benchmark can boost the development of 3D scene understanding, bringing sound reasoning, improved robustness, and solid multi-modal interactions.
	
	{{\section*{Acknowledgment}} This work was supported in part by JCYJ20220530143600001, by the Basic Research Project No. HZQB-KCZYZ-2021067 of Hetao Shenzhen HK S\&T Cooperation Zone, by the National Key R\&D Program of China with grant No.2018YFB1800800, by SGDX20211123112401002, by Shenzhen Outstanding Talents Training Fund, by Guangdong Research Project No. 2017ZT07X152 and No. 2019CX01X104, by the Guangdong Provincial Key Laboratory of Future Networks of Intelligence (Grant No. 2022B1212010001), by the NSFC 61931024\&8192 2046, by NSFC-Youth 62106154, by zelixir biotechnology company Fund, by Tencent Open Fund, and by ITSO at CUHKSZ.}

	
	%

	


	\ifCLASSOPTIONcaptionsoff
	\newpage
	\fi

	
	{
		\bibliographystyle{IEEEtran}
		\bibliography{paper.bbl}
	}

\end{document}